\documentclass{llncs}
\usepackage{makeidx}  % allows for indexgeneration
\usepackage{url}      

\usepackage{graphicx}
\usepackage{amsfonts}
\usepackage{amsmath}
\usepackage{algorithm}
\usepackage{algpseudocode}
\usepackage{subfigure}
\usepackage{multirow}
\usepackage{hyperref}
\usepackage{booktabs}
\usepackage{color}
\usepackage{textcomp}
\usepackage{caption}

\DeclareMathOperator*{\argmin}{\mathrm{arg\:min}}

\begin{document}

\title{X-ray In-Depth Decomposition: Revealing The Latent Structures}
\author{Shadi Albarqouni\inst{1}\and Javad Fotouhi\inst{2}\and Nassir Navab\inst{1,2}}
\institute{Computer Aided Medical Procedures (CAMP), Technische Universit\"at M\"unchen, Munich, Germany
\and 
Whiting School of Engineering, Johns Hopkins University, Baltimore, USA \\
\href{mailto:shadi.albarqouni@tum.de}{shadi.albarqouni@tum.de}}

\maketitle

\begin{abstract}
	X-ray radiography is the most readily available imaging modality and has a broad range of applications that spans from diagnosis to intra-operative guidance in cardiac, orthopedics, and trauma procedures. Proper interpretation of the hidden and obscured anatomy in X-ray images remains a challenge and often requires high radiation dose and imaging from several perspectives. 
	In this work, we aim at decomposing the conventional X-ray image into $d$ X-ray components of independent, non-overlapped, clipped sub-volumes using a deep learning approach. Despite the challenging aspects of modelling such a highly ill-posed problem, exciting and encouraging results are obtained paving the path for further contributions in this direction.
\end{abstract}

\section{Introduction}
\label{sec:intro}
Since its discovery by R\"ontgen in 1895, X-ray is still considered the most accessible imaging modality for both diagnostic and interventional radiology. Its technology is based on the same fundamental principles; emitted X-rays are highly absorbed by hard tissues (\emph{i.e.} bones) leaving the soft tissues with small amount of energy.  
Standard X-ray radiography only produces 2D X-ray images that lacks 3D depth information. Hence, correct interpretation of 3D complex anatomy from a single 2D radiograph remains a challenge.

\begin{figure}[t]
	\centering
	\subfigure[]{\includegraphics[width =0.24\textwidth]{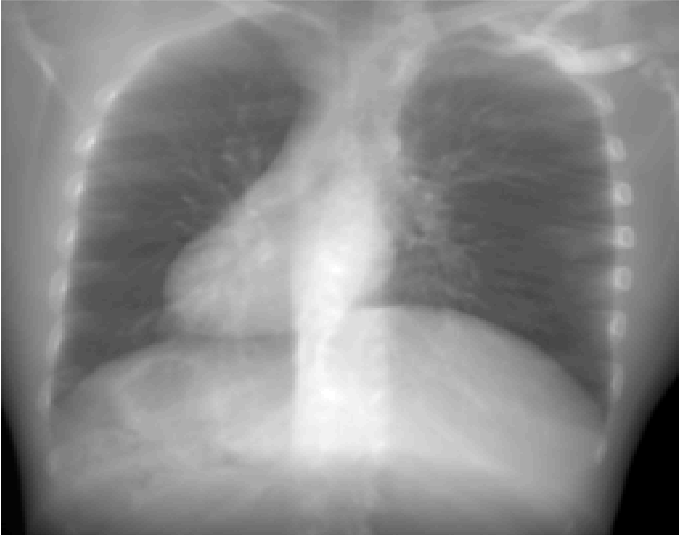}} 
	\subfigure[]{\includegraphics[width =0.24\textwidth]{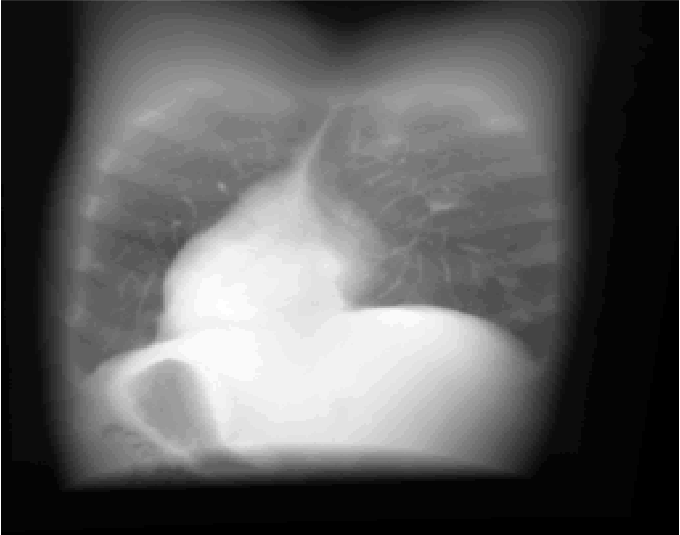}}
	\subfigure[]{\includegraphics[width =0.24\textwidth]{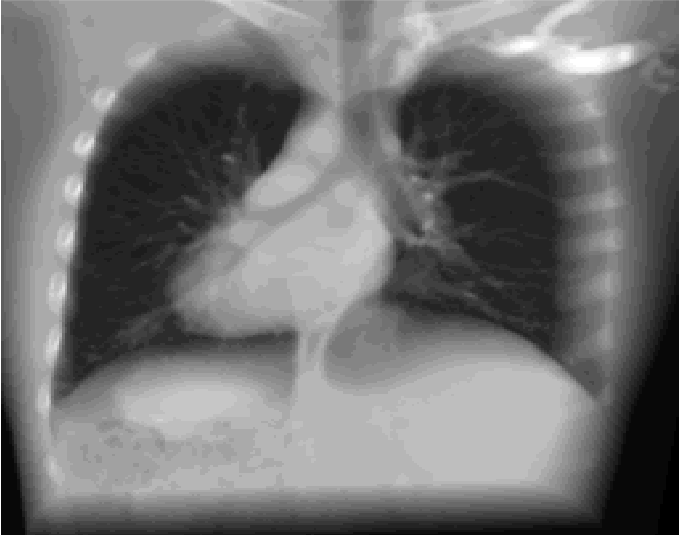}}
	\subfigure[]{\includegraphics[width =0.24\textwidth]{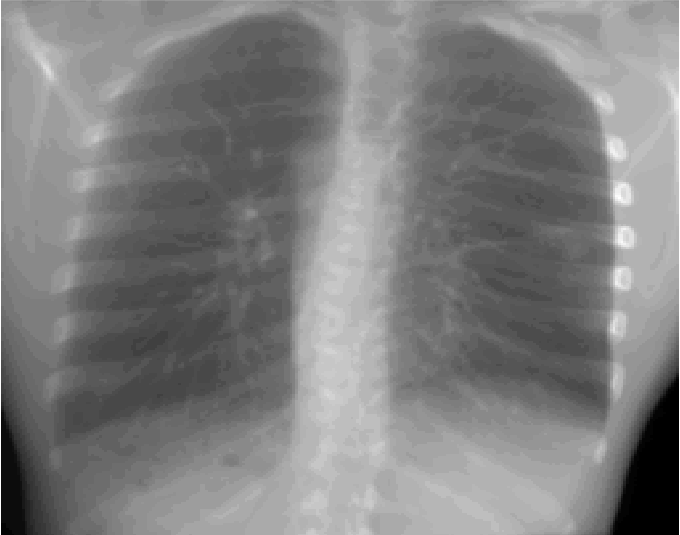}}

	\subfigure[]{\includegraphics[width =0.24\textwidth]{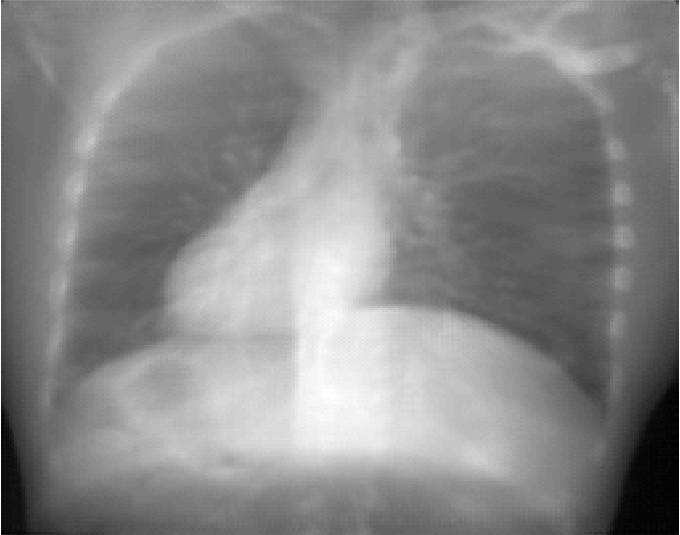}}	
	\subfigure[]{\includegraphics[width =0.24\textwidth]{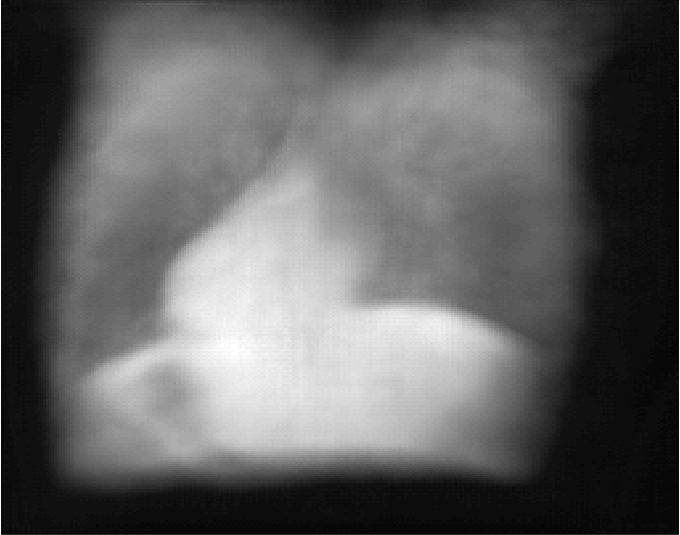}}
	\subfigure[]{\includegraphics[width =0.24\textwidth]{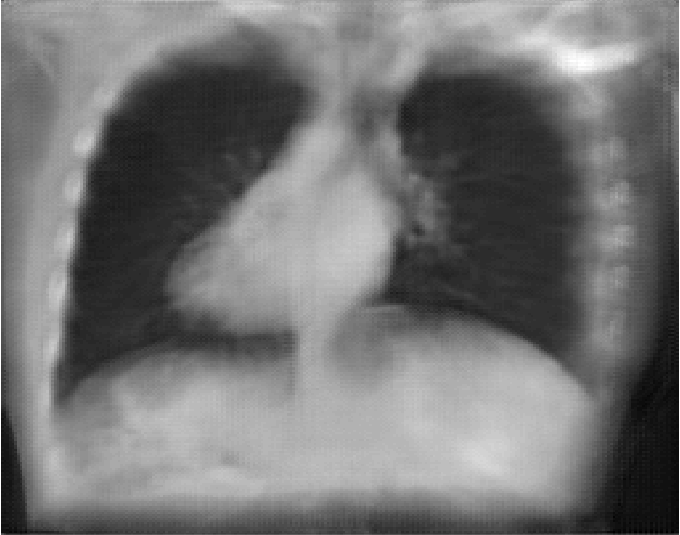}}
	\subfigure[]{\includegraphics[width =0.24\textwidth]{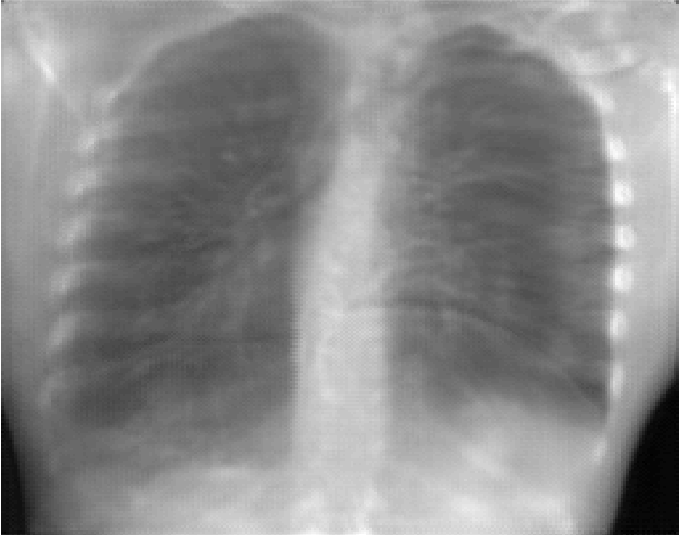}}
	\caption{Simulated ground-truth X-ray projections of (a) Thorax $\vec{\mu_T}$, (b) Ribcage  $\vec{\mu_R}$, (c) Vasculature $\vec{\mu_V}$, and (d) Spine $\vec{\mu_S}$ sub-volumes. Our in-depth X-ray decompositions (f-h). The total sum of decomposed layers (e) resembles the X-ray projection in (a).}
	\label{fig:XrayDecomposition}
	\vspace{-0.2cm}
\end{figure}

In diagnostic settings, observing chest X-ray (CXR) and distinguishing the anterior and posterior ribs, spine, and more importantly soft tissues, \emph{i.e.} pulmonary vascular tree, requires skilled clinicians (see Fig~\ref{fig:XrayDecomposition}) and heavily relies on their perceptual skills and judgment~\cite{Brady2017}. This holds for surgical settings as well, when looking at interventional X-rays, the most experienced radiologists and vascular surgeons  can focus on anatomy of interest and ignore the surrounding rigid anatomy such as rib cage and spine. In some sense, they do see through consistent layers of known anatomy. The main question, we are trying to answer is whether a computer can do the same. Can the computer learn to separate different layers of anatomy and let us focus on a given layer in which there is high variation and is of interest for a given clinical decision. We know that this is mathematically an ill-posed position. For many years mathematicians looked at such problems and the need for multi-view acquisition or well-defined priori knowledge is fully established.
Recent advances in visualization and augmented reality allow using prior knowledge from pre-operative patient CT data to visualize depth information overlaid on the medical data~\cite{kersten2006enhancing,wieczorek2011interactive,wang2014augmented}. However, accurate registration between 3D pre-operative CT scan and 2D X-ray image is required, which is a complex problem, and in several cases disrupts the surgical or clinical workflow. However, the question remains whether such a prior knowledge could be fully learned through a well learning strategy and a well designed machine learning tool? 

In this paper, we take this challenge and aim at decomposing a single X-ray acquired for a given clinical procedure and learn how to decompose it into it is predefined set of layers. It is important to note that we are not aiming at the general problem of decomposition of any X-ray image into layers as we consider it highly ill-posed and mathematically impossible. We choose the clinical problem such that some of the layers have morphological properties that allow for the learning algorithm to discover, and only one layer have high variation and would be hard to learn. The additive property of X-ray absorption physics allows the algorithm to recover the correct projection of such a complex layer, as it is in fact complementary to the sum of absorptions of other layers forming together the input X-ray image.  There are many applications for such decomposition and probably the first and most important one is the reduction of noise and improvement of the visibility of structures of interest in given layers of interest for particular diagnosis or intervention. To our knowledge, this is the first work modeling the problem as such and aiming at solving it with the most advance tools in machine learning. 

\section{Methodology}
\label{sec:method}
Our methodology, depicted in Fig.~\ref{fig:framework}, utilizes a pre-operative CT scan, which is commonly acquired two weeks prior to the intervention, where the clipping planes are defined by the clinicians. To generate the training data, simulated X-ray images, so-called Digitally Reconstructed Radiographs (DRRs), are generated along trajectories which fully captures the region of interest (ROI) \emph{i.e.}~Abdominal Aortic Aneurysm (AAA). 
This is done by first slicing the patient CT volume in the anterior-posterior plane to several layers (clipping planes). DRRs are generated from different perspectives of the sub-volumes as well as the original CT data. 
These DRRs form the training set of our model. In the testing phase, for any given X-ray image, decomposition of $d$ X-ray components is obtained. %\todo{reorder some figures}

\begin{figure}[t]
	\centering 
	\includegraphics[height= 0.3\textheight,width = \textwidth]{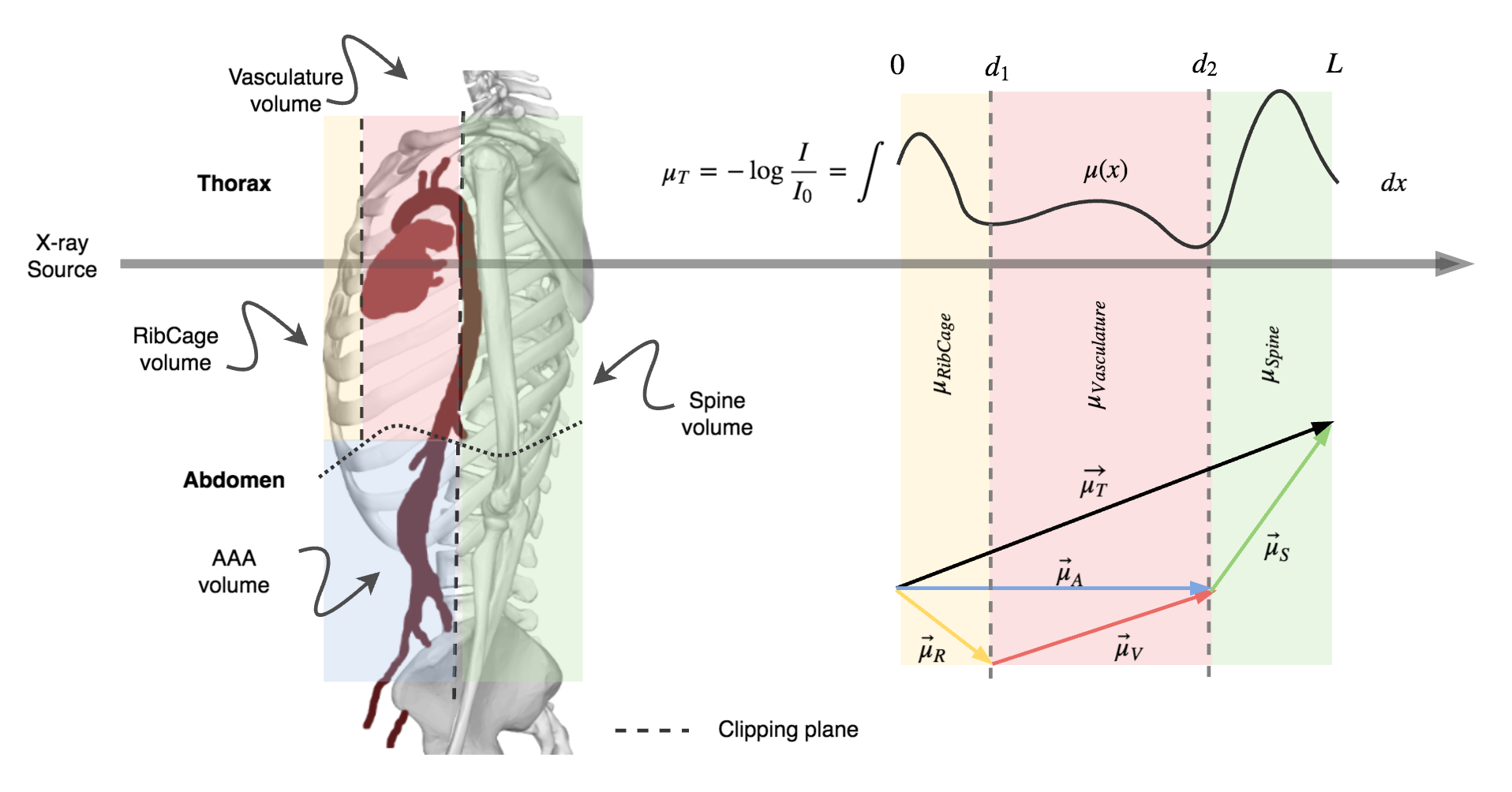}
	\caption{Illustrative diagram of clipped CT volume (left) and X-ray decomposition in vector space (right).}
	\label{fig:VectorSum}
\end{figure}

\subsection{Problem Formulation}
Based on Beer-Lambert Law, the absorbance $\vec{\mu_{T}}$ of a given X-ray image is formed by accumulating the Hounsfield Units (HU) of a CT volume $\mu(x)$ along the ray cast. 
In an example case of a CT volume sliced into three sub-volumes, following Beer-Lambert Law, the absorbance can be written as:
\begin{equation}
	\label{eq:logBeer}
\vec{\mu_{T}} = - \log{\left(\frac{I}{I_0}\right)} = \int_{0}^{d_1} \! \mu(x) \, dx + \int_{d_1}^{d_2} \! \mu(x) \, dx + \int_{d_2}^{L} \! \mu(x) \, dx
\end{equation}
where $I$ and $I_0$ are the incident and maximal radiation respectively, and $d_1$ and $d_2$ are clipping planes (see Fig.~\ref{fig:VectorSum}). This can be re-written as a linear sum of X-ray projections of these clipped sub-volumes, 
\begin{equation}
\label{eq:Decomposition}
\vec{\mu_{T}} = \sum_{d} \vec{\mu_d} = \vec{\mu_R}  + \vec{\mu_V} + \vec{\mu_S}, 
\end{equation}
where $\vec{\mu_{T}}, \vec{\mu_R}, \vec{\mu_V}, \text{and~} \vec{\mu_S} \in \mathbb{R}^{H \times W}$ are the corresponding X-ray images of chest (thorax), ribcage, vasculature and spine sub-volumes, respectively. 
Our objective is that for any given X-ray image $\vec{\mu_T}$, the proposed model $f(\cdot)$ predicts $d$ independent X-ray components $\vec{\hat{\mu_d}}$, where the total sum of these components is equal to the original image: 
\begin{equation}
\label{eq:AbdomenFomrula}
\vec{\hat{\mu_d}} = \{\vec{\hat{\mu_R}}, \vec{\hat{\mu_V}}, \vec{\hat{\mu_S}} \} = f(\vec{\mu_T}; w), \hspace{0.5cm} \emph{s.t.} \hspace{0.3cm} \sum_d \vec{\hat{\mu_d}} = \vec{\hat{\mu_R}} + \vec{\hat{\mu_V}}  + \vec{\hat{\mu_S}} = \vec{\mu_{T}},
\end{equation}
where $\vec{\hat{\mu_d}} \in \mathbb{R}^{H \times W \times d}$
is a stack of the predicted outputs and $w$ is the model parameters. This objective function can be formulated using Lagrangian multiplier as:
\begin{equation}
\label{eq:AbdomenObjective}
\argmin_{\vec{\hat{\mu_d}}} \hspace{0.2cm} \underbrace{\mathcal{L}_d\left(\vec{\hat{\mu_d}}, \vec{\mu_d} \right)}_{decomposition} + \lambda_r \underbrace{\mathcal{L}_{r}(\vec{\hat{\mu_d}}, \vec{\mu_T})}_{reconstruction}
\end{equation}
where $\lambda_r$ is the regularization parameter, and $\mathcal{L}_d(\cdot,\cdot)$ is the elastic-net loss
between the predicted decompositions and their corresponding ground-truths:
\begin{equation}
\label{eq:elastic-net}
\mathcal{L}_d\left(\vec{\hat{\mu_d}}, \vec{\mu_d} \right) = \lambda_d \underbrace{\| \vec{\hat{\mu_d}} - \vec{\mu_d} \|_2}_{smoothness} + (1-\lambda_d) \underbrace{\| \vec{\hat{\mu_d}} - \vec{\mu_d} \|_1}_{sparsity},
\end{equation}
where $\lambda_d$ controls the tradeoff between $\ell_1$ and $\ell_2$ norms. Finally, $\mathcal{L}_r(\cdot,\cdot)$ is the $\ell_2$-loss between the sum of predicted outputs and the given input:
\begin{equation}
\label{eq:reconstruction}
\mathcal{L}_{r}\left(\vec{\hat{\mu_d}}, \vec{\mu_T} \right) = \| \sum_d \vec{\hat{\mu_d}} - \vec{\mu_T} \|_2
\end{equation}

DRRs are generated from the original CT volume (stored in $\vec{\mu_T}$) as well as the sub-volumes (stored in $\vec{\mu_d}$) with fixed intrinsic parameters (estimated from the C-arm calibration), and different extrinsic parameters which simulate realistic trajectories of a C-arm.  

\begin{figure}[t]
	\centering 
	\includegraphics[width = \textwidth]{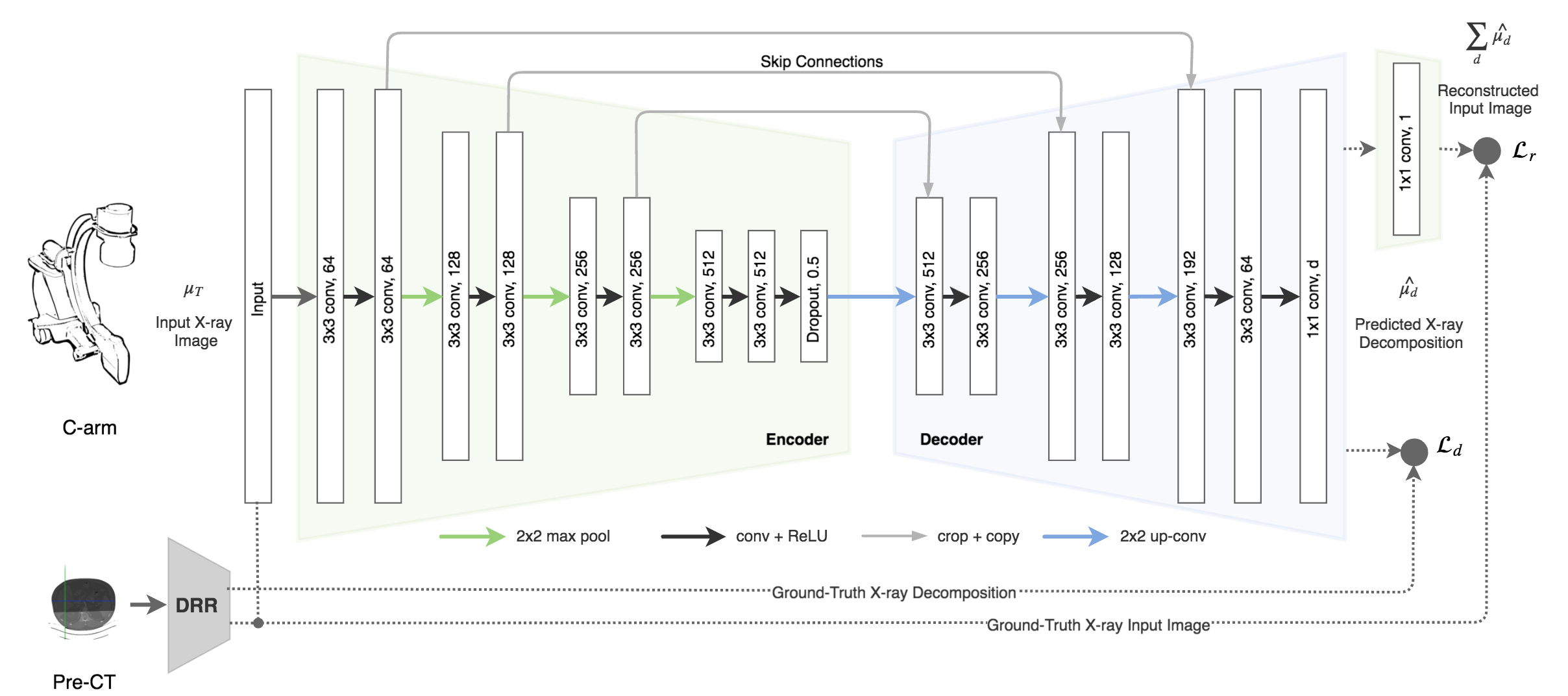}
	\caption{X-ray decomposition framework: In the training phase (dotted lines), DRRs are generated from pre-operative CT scans and their corresponding clipped sub-volumes, while in testing phase, an X-ray image is acquired using a C-arm and fed to the network to output the corresponding X-ray decompositions.}
	\label{fig:framework}
	\vspace{-0.2cm}
\end{figure}

\paragraph{Network Architecture:}
Our network architecture (cf. Fig.~\ref{fig:framework}) is similar to U-Net~\cite{ronneberger2015u} with slight modifications, \emph{i.e.}~the last convolutional layers were dropped, and a dropout is added immediately after the last convolutional layer in the encoder to avoid minor overfitting and speed up the training process. In addition, we incorporate decomposition loss $\mathcal{L}_d$, which employs elastic-net loss, right after the last convolutional layer in the decoder part. 
To fulfill the reconstruction constraint (Beer-Lambert law), a $1 \times 1$ convolutional layer is added right before the reconstruction loss $\mathcal{L}_r$.

\section{Experiments and Results}
\label{sec:expresults}
We designed two experimental setups to validate our proposed methodology. While the first experiment is designed for application-specific model; i) intra-operative and ii) diagnostic purposes, the second experiment targets a clinical usecase. Models that are trained without the reconstruction loss ($\lambda_r =0$) are considered as baseline models.
\paragraph{Dataset:}
We validated our proposed method on 6 clinical datasets 
acquired with different protocols for thorax and abdomen. 
For each clinical dataset, approximately 1200 ground-truth DRRs are rendered for the whole volume as well as the corresponding sub-volumes, each of size $256 \times 256$ pixels. DRRs are divided into non-overlapping sets of training ($60\%$), validation ($20\%$), and testing ($20\%$). Patient-wise splitting is considered for the diagnostic setup, \emph{i.e.}~training ($67\%$) and testing ($33\%$). Furthermore, to validate our methodology for a clinical usecase, we opt an additional database of chest X-ray (615 patients) for Tuberculosis (TB) classification~\cite{jaeger2014automatic}.

\paragraph{Simulated X-ray of Clipped CT Volume:}
\label{DRRs}As depicted in Fig.~\ref{fig:VectorSum}, the pre-operative CT is clipped at two coronal planes of thorax (one coronal plane for abdomen) resulting in three sub-volumes; ribcage, vasculature, and spine (two sub-volumes; AAA and spine for abdomen). These sub-volumes are padded with background values to retain the same dimension of the original CT. A C-arm simulator software (ImFusion\footnote{\url{http://imfusion.de/products/imfusion-suite}}) is used to generate DRRs along pre-defined trajectories, \emph{i.e}.~Cranial ($0-20^\circ$), Left Anterior Oblique (LAO) and Right Anterior Oblique (RAO) ($0-40^\circ$), for both original volume and different sub-volumes as well.

\paragraph{Implementation:}
The learning rate was set to $10^{-6}$, momentum to $0.9$ and batch size to $16$. The hyper-parameters $\lambda_r$ was set to $0.5$, and $\lambda_d$ was altered between $0.1$ and $0.9$ to observe the influence of each term. 
\paragraph{Evaluation:}
Two different metrics are computed to evaluate the performance of our model; PSNR and Structural Similarity Index (SSIM) to measure the quality of reconstruction and perception, respectively. 

\subsection{Application-Specific Model}
To address more realistic scenarios, we designed two experiments for: \emph{i)} intra-operative scenarios, where the model is trained on several pre-operative patient CT data, and tested on the same patient data during the intervention, and \emph{ii)} diagnostic applications which rely solely on X-ray image (CT not needed).

\paragraph{Intra-operative Purpose:}
In this setup, a pre-operative CT scan is required for a new patient, and is used to further fine-tune the pre-trained model on previous patients. A model is trained on three patients (2160 images) for 200 epochs and tested on the same patients for entirely different trajectories (720 images). Results on testing sets, per patient and overall, are reported for both Abdomen and Thorax datasets in Tables~\ref{tab:AbdomenIntra} and \ref{tab:ThoraxIntra}, respectively. A comparison with baseline models ($\lambda_r =0$) shows a positive influence of the reconstruction loss on the desired decompositions in terms of better PSNR and SSIM.

\begin{table}[t]
	\centering
			\begin{minipage}[t]{0.49\hsize}\centering
				\caption{Intra-op. purpose (abdomen)}
					\resizebox{\textwidth}{!}{
						\label{tab:AbdomenIntra}
						\begin{tabular}{c|c|c|c|c|c|c|c}			
							\toprule
							&\multirow{2}{*}{Patient \textnumero} & \multicolumn{3}{c}{PSNR (dB)} \vline & \multicolumn{3}{c}{SSIM (100\%)}\\
							\cmidrule{3-8}	
							&    & AAA & Spine & Abdomen & AAA & Spine & Abdomen\\
							\hline
							\parbox[t]{4mm}{\multirow{4}{*}{\rotatebox[origin=c]{90}{$\lambda_r = 0.5$}}} &25 & 53.98  & 51.31 & 23.71  & 83.69  & 83.46  & 82.98 \\
							&36 & 56.08  & 55.31  & 27.08  & 87.22  & 86.17 & 81.96 \\			
							&37 & 53.98  & 49.43 & 27.70  & 87.52  & 82.08  & 81.00 \\
							\cmidrule{2-8}
							&Overall & \textbf{53.02}  & \textbf{50.62} & 26.00 & \textbf{86.14}  & \textbf{83.90}  & \textbf{81.98}\\
							\hline
							\parbox[t]{4mm}{\multirow{4}{*}{\rotatebox[origin=c]{90}{$\lambda_r = 0$}}} &25 & 51.56 & 50.89 & 23.58  & 81.05  & 83.02  & 82.16 \\
							&36 & 55.59 & 52.81 & 27.21  & 86.08  & 85.95  & 80.81  \\			
							&37 & 52.87 & 48.79 & 27.96  & 86.99  & 82.04  & 80.10 \\
							\cmidrule{2-8}
							&Overall & 52.06 & 49.77 & \textbf{26.05}  & 84.70  & 83.67  & 81.02 \\			
							\bottomrule
						\end{tabular}
					}						
			\end{minipage}
			\begin{minipage}[t]{0.49\hsize}\centering
			\caption{LOPOCV (abdomen)}
					\resizebox{\textwidth}{!}{
						\label{tab:AbdomenLOPOCV}
						\begin{tabular}{c|c|c|c|c|c|c|c}	
							\toprule
							&Patient \textnumero & \multicolumn{3}{c}{PSNR (dB)} \vline & \multicolumn{3}{c}{SSIM (100\%)}\\
							\cmidrule{2-8}	
							&train (test)& AAA & Spine & Abdomen & AAA & Spine & Abdomen\\
							\hline
							\parbox[t]{4mm}{\multirow{4}{*}{\rotatebox[origin=c]{90}{$\lambda_d = 0.1$}}}&36,37 (25) & 45.78  & 46.55 & 24.81  & 82.24  & 83.75  & 83.58 \\
							&25,37 (36) & 50.78  & 50.40 & 25.08  & 83.76  & 83.21  & 79.54 \\			
							&25,36 (37) & 49.19  & 41.59 & 29.98  & 85.65  & 80.04  & 82.72 \\
							\cmidrule{2-8}
							&Overall & 48.36  & \textbf{45.52} & 26.36  & 83.89  & \textbf{82.33}  & 81.95   \\
							\hline
							\parbox[t]{4mm}{\multirow{4}{*}{\rotatebox[origin=c]{90}{$\lambda_d = 0.9$}}}&36,37 (25) & 45.84  & 46.56 & 24.78  & 82.26  & 83.81  & 83.73 \\
							&25,37 (36) & 52.37  & 47.75 & 26.36  & 84.65  & 84.33  & 80.66 \\			
							&25,36 (37) & 48.15  & 40.21 & 30.85  & 85.60  & 77.80  & 83.32 \\
							\cmidrule{2-8}
							&Overall & \textbf{48.44}  & 44.26 & \textbf{27.02}  & \textbf{84.17}  & 81.98  & \textbf{82.57}  \\
							\bottomrule
						\end{tabular}
					}
			\end{minipage}
\end{table}

\begin{table}[t]
		\caption{Intra-operative purpose (thorax)}
		\label{tab:ThoraxIntra}
		\begin{tabular*}{\textwidth}{@{}c|c|c|c|c|c|c|c|c|c@{}}
			\toprule
			&\multirow{2}{*}{Patient \textnumero} & \multicolumn{4}{c}{PSNR (dB)} \vline & \multicolumn{4}{c}{SSIM (100\%)}\\
			\cmidrule{3-10}	
			& & Ribcage & Vasculature & Spine & Thorax & Ribcage & Vasculature & Spine & Thorax\\
			\hline
			\parbox[t]{4mm}{\multirow{4}{*}{\rotatebox[origin=c]{90}{$\lambda_r = 0.5$}}}& 1 & 83.97  & 88.72  & 73.21  & 54.60  & 90.96  & 96.67  & 96.14  & 92.60 \\
			
			& 2 & 74.74 & 84.54  & 78.85 & 57.21  & 89.54  & 93.22  & 95.42  & 92.88 \\
			
			& 4 & 83.45 & 91.39  & 84.65  & 55.70  & 92.71  & 97.20  & 97.73 & 93.32 \\
			\cmidrule{2-10}	
			& Overall & \textbf{78.97} & 87.36  & \textbf{77.48}  & \textbf{55.77}  & \textbf{91.07}  & \textbf{95.70}  & \textbf{96.43}  & \textbf{92.93}\\
			\hline
			\parbox[t]{4mm}{\multirow{4}{*}{\rotatebox[origin=c]{90}{$\lambda_r = 0$}}}& 1 & 70.41  & 80.30  & 65.56  & 54.45  & 80.55  & 94.23  & 93.34  & 91.63 \\
			
			& 2 & 68.84 & 81.22  & 73.29 & 55.06  & 81.47  & 88.68  & 93.45  & 90.51 \\
			
			& 4 & 72.77 & 87.18  & 76.23  & 54.03  & 85.98  & 94.43  & 95.78 & 91.59 \\
			\cmidrule{2-10}	
			& Overall & 75.16 & \textbf{87.44}  & 74.28  & 54.83  & 87.73  & 95.12  & 95.49  & 91.49\\			
			\bottomrule
		\end{tabular*}
	\vspace{-10pt}
\end{table}

\paragraph{Diagnostic Purpose:}
In this setup, a pre-operative CT scan is not required for testing.  The model is trained on two patients (2400 images) for 60 epochs ($\lambda_r = 0.5$), and tested on the third patient (1200 images). Leave One Patient Out Cross Validation (LOPOCV) on both Abdomen and Thorax datasets (cf. Tables~\ref{tab:AbdomenLOPOCV} and \ref{tab:thoraxLOPOCV}) was performed to investigate the robustness and validate the influence of hyper-parameter $\lambda_d$. For instance, the high contribution of sparsity term $\ell_1$ gives better results on the thorax-spine image (SSIM=93.57) than the same contribution of smoothness term $\ell_2$ (SSIM=91.90).    

\begin{table}[t]
		\caption{LOPOCV for diagnostic purpose (thorax)}
		\label{tab:thoraxLOPOCV}
		\centering
		\begin{tabular*}{\textwidth}{@{}c|c|c|c|c|c|c|c|c|c@{}}
			
			\toprule
			&Patient \textnumero & \multicolumn{4}{c}{PSNR (dB)} \vline & \multicolumn{4}{c}{SSIM (100\%)}\\
			\cmidrule{2-10}	
			&train (test) & Ribcage & Vasculature & Spine & Thorax & Ribcage & Vasculature & Spine & Thorax\\
			\hline
			 \parbox[t]{4mm}{\multirow{4}{*}{\rotatebox[origin=c]{90}{$\lambda_d = 0.1$}}} & 2,4 (1) & 73.23  & 80.28  & 61.70  & 60.62  & 83.19  & 95.53  & 90.83  & 95.16  \\
			
			&1,4 (2) & 71.15 & 85.15  & 72.36 & 54.65  & 84.06  & 92.16  & 92.96  & 90.10\\
			
			&1,2 (4) & 78.19 & 92.42  & 78.21  & 53.79  & 89.13  & 96.61  & 96.93 & 92.09 \\
			\cmidrule{2-10}
			&Overall & \textbf{73.78} & \textbf{84.78}  & \textbf{68.40}  & \textbf{55.93}  & 85.46  & \textbf{94.10}  & \textbf{93.57}  & \textbf{92.45}\\
			\hline
			 \parbox[t]{4mm}{\multirow{4}{*}{\rotatebox[origin=c]{90}{$\lambda_d = 0.9$}}} & 2,4 (1) & 73.80  & 82.29  & 61.72  & 60.67  & 86.16  & 93.94  & 90.46  & 95.53 \\
			 
			 &1,4 (2) & 67.48 & 84.48  & 66.01 & 54.30  & 84.38  & 92.44  & 90.88  & 90.49\\
			 
			 &1,2 (4) & 75.65 & 87.22  & 70.52  & 52.27  & 86.13  & 94.79  & 93.46 & 90.85\\
			 \cmidrule{2-10}
			 &Overall & 71.67 & 84.46  & 65.45  & 55.16  & \textbf{85.56}  & 93.72  & 91.90  & 92.29\\
			\bottomrule
		\end{tabular*}
	\vspace{-10pt}
\end{table}

\subsection{Clinical Usecase}
\begin{figure}[t]
	\centering
	\begin{minipage}{0.55\textwidth}
		\subfigure{\includegraphics[width=0.32\textwidth]{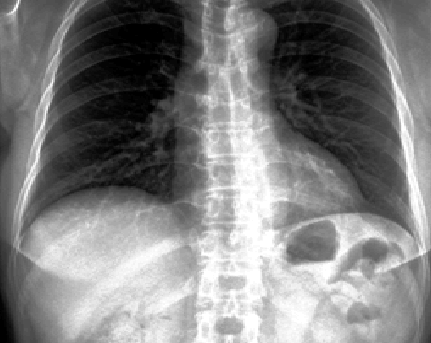}}
		\subfigure{\includegraphics[width=0.32\textwidth]{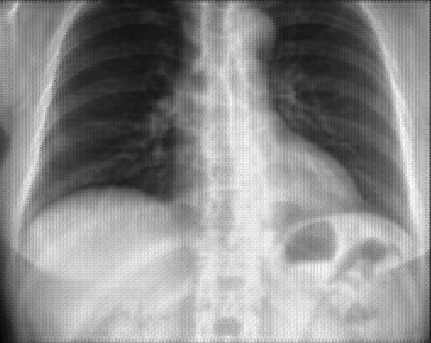}}
		\subfigure{\includegraphics[width=0.32\textwidth]{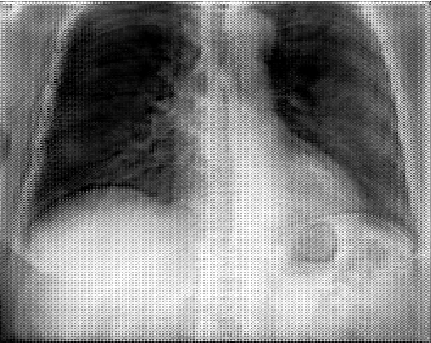}}
	
		\subfigure{\includegraphics[width=0.32\textwidth]{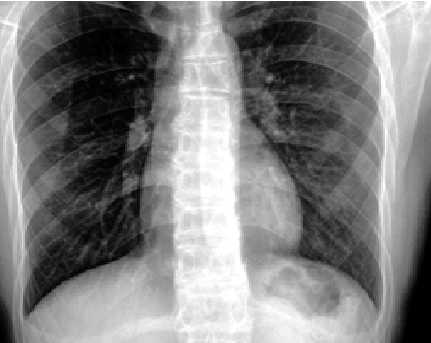}}
		\subfigure{\includegraphics[width=0.32\textwidth]{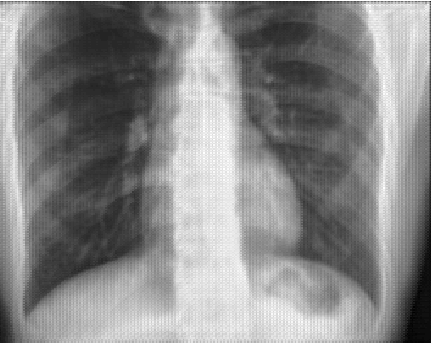}}
		\subfigure{\includegraphics[width=0.32\textwidth]{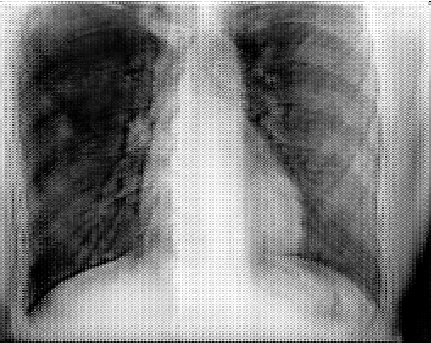}}
		\caption{Real CXR (left), reconstructed CXR (middle), vasculature decomposition (right).}
		\label{fig:TBdempo}
	\end{minipage}
	\begin{minipage}{0.44\textwidth}
		\centering
		\subfigure{\includegraphics[width=0.9\textwidth]{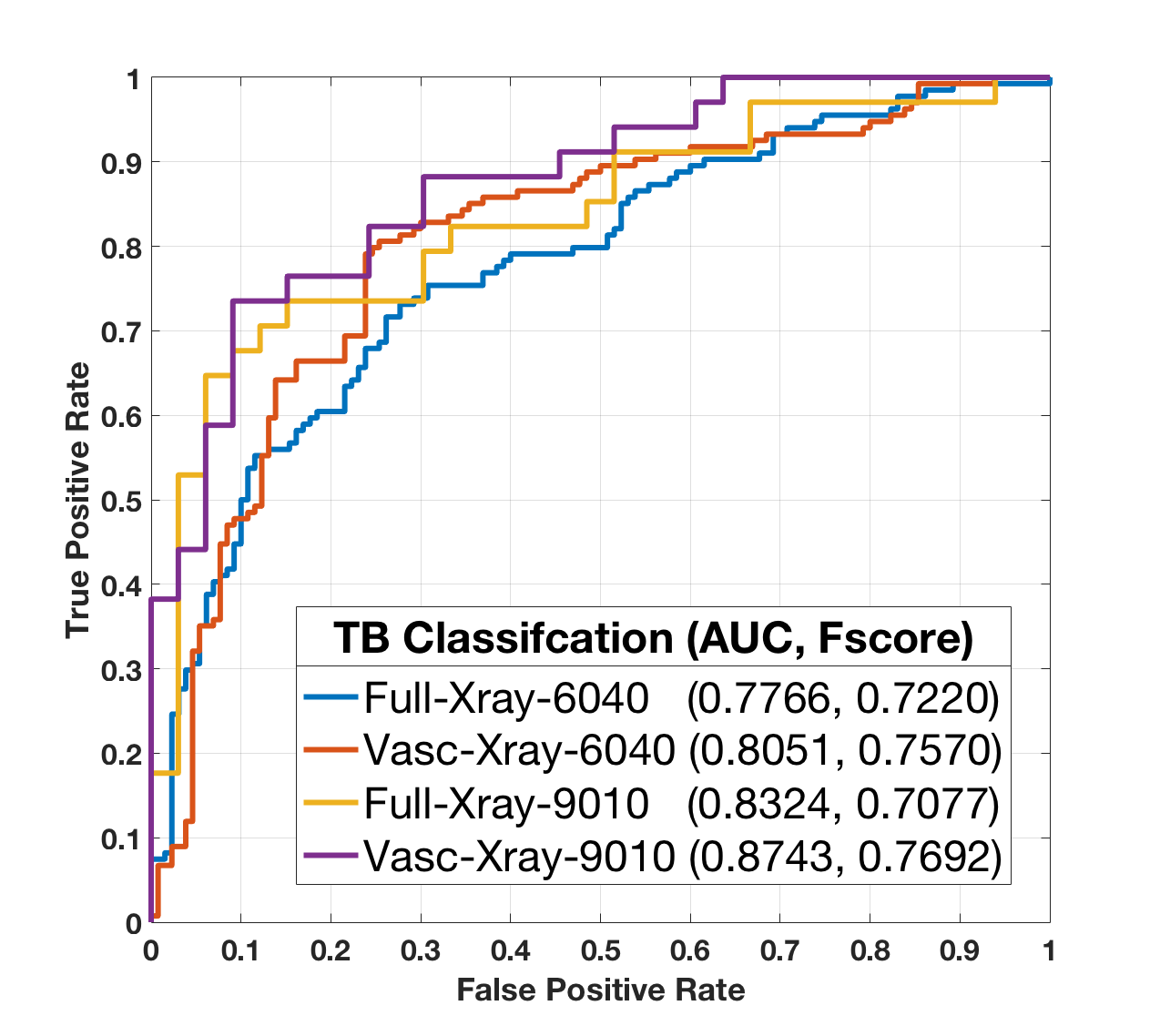}}
		\caption{ROC curves}
		\label{fig:TBclass}
	\end{minipage}
	\vspace{-10pt}
\end{figure}

In this experiment, we designed a CAD model for automatic TB classification, based on ResNet-16 architecture~\cite{he2016deep}, to present a clinical usecase for the proposed methodology. 615 real CXR images, forming \verb|Full_Xray| set, went through our previous model (intr-operative purpose) to estimate the X-ray decompositions. The vasculature components are then collected to form \verb|Vasc_Xray| set (cf. Fig.~\ref{fig:TBdempo}). In this experiment, we considered different splittings of training and testing sets, \emph{i.e.} 90:10\% and 60:40\%. We observed that models trained on vasculature X-ray components yield significant improvement of AUC and $F_1score$ compared to the models trained from original images (cf. Fig.~\ref{fig:TBclass}).

\section{Discussion and Conclusion}
\label{sec:conclusion}
This work presents a novel methodology that uses deep learning for in-depth decomposition of single-view X-ray images.
As anticipated, the intra-operative models perform better than the diagnostic ones due to the heterogeneity in the patient population. We also observed that our proposed model captured the rigid anatomical structures efficiently, \emph{i.e.} ribs and spine, while it did not perform well for deformable ones (see Fig.~\ref{fig:XrayDecomposition}). This justifies the modest number of medical datasets. Large amount of data (including real X-ray images) to model the heterogeneity available in real scenarios (such as obesity and gender) should be further investigated. 
TB classification results confirm our hypothesis that our in-depth decomposition reveals latent structures that improve the perception and clinical interpretation.
Yet, additional clinical validation is required for diagnostic and interventional setups.
\bibliography{XrayDepth,biblio-macros}

\begin{thebibliography}{1}
\providecommand{\url}[1]{\texttt{#1}}
\providecommand{\urlprefix}{URL }

\bibitem{Brady2017}
Brady, A.P.: Error and discrepancy in radiology: inevitable or avoidable?
  Insights into Imaging  8(1),  171--182 (2017)

\bibitem{he2016deep}
He, K., Zhang, X., Ren, S., Sun, J.: Deep residual learning for image
  recognition. In: Proceedings of the IEEE on CVPR. pp. 770--778 (2016)

\bibitem{jaeger2014automatic}
Jaeger, S., Karargyris, A., Candemir, S., Folio, L., Siegelman, J., Callaghan,
  F., Xue, Z., Palaniappan, K., Singh, R.K., Antani, S., et~al.: Automatic
  tuberculosis screening using chest radiographs. IEEE TMI  33(2),  233--245
  (2014)

\bibitem{kersten2006enhancing}
Kersten, M., Stewart, J., Troje, N., Ellis, R.: Enhancing depth perception in
  translucent volumes. IEEE TVCG  12(5) (2006)

\bibitem{ronneberger2015u}
Ronneberger, O., Fischer, P., Brox, T.: U-net: Convolutional networks for
  biomedical image segmentation. In: MICCAI. pp. 234--241. Springer (2015)

\bibitem{wang2014augmented}
Wang, J., Kreiser, M., Wang, L., Navab, N., Fallavollita, P.: Augmented depth
  perception visualization in 2d/3d image fusion. Computerized Medical Imaging
  and Graphics  38(8),  744--752 (2014)

\bibitem{wieczorek2011interactive}
Wieczorek, M., Aichert, A., Fallavollita, P., Kutter, O., Ahmadi, A., Wang, L.,
  Navab, N.: Interactive 3d visualization of a single-view x-ray image. In:
  MICCAI. pp. 73--80. Springer (2011)

\end{thebibliography}
\bibliographystyle{splncs03}

\end{document}